\documentclass[letterpaper, 10 pt, conference]{ieeeconf}  

\IEEEoverridecommandlockouts                              

\usepackage{color}
\usepackage{hyperref}

\newcounter{mnotei}
\title{Towards the Safety of Human-in-the-Loop Robotics:\\ Challenges
  and Opportunities for Safety Assurance of Robotic Co-Workers$^*$}

\author{Kerstin Eder$^{1}$, Chris Harper$^{2}$ and Ute
  Leonards$^{3}$
  \thanks{*The authors were brought together during a Safe and
    Trustworthy Autonomous Assistive Robots (STAARs) workshop
    sponsored by the Institute for Advanced Studies at the University
    of Bristol. 
  }
  \thanks{$^{1}$Kerstin Eder is with the Department of Computer
    Science, University of Bristol, UK, and leads the Verification and
    Validation for Safety in Robots research theme at the Bristol
    Robotics Laboratory, Bristol, UK. {\tt\small
      Kerstin.Eder@bristol.ac.uk}}%
  \thanks{$^{2}$Chris Harper is with Avian Technologies Ltd and a
    Visiting Fellow at the Bristol Robotics Laboratory, Bristol, UK.
    {\tt\small cjharper@avian-technologies.co.uk}}%
  \thanks{$^{3}$Ute Leonards is with the School of Experimental
    Psychology, University of Bristol, UK, and leads the Safe Human
    Robot Interaction research theme at the Bristol Robotics
    Laboratory, Bristol, UK. {\tt\small Ute.Leonards@bristol.ac.uk}}%
}

\begin{document}

\maketitle
\thispagestyle{empty}
\pagestyle{empty}

\begin{abstract}
  The success of the human-robot co-worker team in a flexible
  manufacturing environment where robots learn from demonstration
  heavily relies on the correct and safe operation of the robot. How
  this can be achieved is a challenge that requires addressing both
  technical as well as human-centric research questions.
  In this paper we discuss the state of the art in safety assurance,
  existing as well as emerging standards in this area, and the need
  for new approaches to safety assurance in the context of learning
  machines.
  We then focus on robotic learning from demonstration, the challenges
  these techniques pose to safety assurance and outline opportunities
  to integrate safety considerations into algorithms ``by design''.
  Finally, from a human-centric perspective, we stipulate that, to
  achieve high levels of safety and ultimately trust, the robotic
  co-worker must meet the innate expectations of the humans it works
  with.
  It is our aim to stimulate a discussion focused on the safety aspects
  of human-in-the-loop robotics, and to foster multidisciplinary
  collaboration to address the research challenges identified.
\end{abstract}

\section{INTRODUCTION}

Robotic co-workers are machines designed to support flexible
manufacturing in collaboration with humans. They complement the skills
and cognitive abilities that enable humans to accomplish frequently
changing, varied or imprecise tasks, with strength, precision,
endurance and limitless capacity for repetition.
These robots are expected to provide assistance for a wide variety of
tasks.
To reduce or even eliminate the effort involved with frequently
re-programming robots so that they can perform new tasks, techniques
that enable robot learning from demonstration are now being
developed~\cite{argall}. Such techniques empower non-experts to teach or
train robots, e.g.\ how to be most useful within a flexible
manufacturing environment.

Flexible manufacturing requires robotic co-workers to act within the
personal space of a human. They may be involved in shared manipulation
of objects and even make direct contact with their human operators. To
be genuinely useful, some robots may need to be powerful and therefore
are potentially dangerous. Safety of the humans who interact with
these machines is clearly a prime concern in these settings.
The introduction of learning from demonstration techniques further
emphasizes the need to ensure the safety of human operators, both
during the learning phase and also afterwards, when the newly acquired
task is being performed by the robot.
The safe operation of the robotic co-workers is a core foundation for
humans to establish trust in them.

In this paper we investigate issues surrounding safety assurance of
robotic co-workers.
%
%
We start from a technical, robot-centric angle by considering safety
assurance and certification requirements currently under development
for robots that work in direct interaction with humans in shared
spaces.
We show that there is a considerable gap to bridge, and hence a
research challenge associated with safety assurance of robots that
flexibly learn new tasks ``on the fly''.  We review proposals to shift
parts of safety assurance from design time to runtime, and investigate
the feasibility of using similar techniques to achieve safety
assurance for robotic co-workers.
We then focus on robot ``learning from demonstration'' techniques, in
particular on reinforcement learning, and illustrate both the
opportunities arising from humans in the loop, as well as the dangers
and the associated responsibility to ensure the safety of the human
operators.
We aim to identify opportunities to integrate safety considerations
into the design of robotic co-workers, i.e.\ into the embedded
learning algorithms from the beginning ``by design''.

Our discussion then takes a human-centric view and focuses on the
psychological challenges associated with human-robot co-worker teams.
We stipulate that, to achieve high levels of safety and ultimately
trust, the robotic co-worker must meet the innate expectations of the
humans it works with.
For instance, if a co-worker robot were to be equipped with head and
eyes, it would be expected to use eye gaze as warning signal in case
of danger by directing gaze toward the source of danger. Also, in a
shared activity, gaze would be expected to be directed to the task
at hand in line with human expectations on the focus of attention.
This requires deep understanding of the signals sent by humans and the
way humans interpret these signals when observed in human co-workers.

With this position paper we aim to stimulate a discussion focused on
safety aspects of robotic co-workers, both from a robot design as well
as from a human-centric perspective. Our paper poses more research
questions than it provides answers.
These questions, however, are designed to open up an opportunity for
the research community to address safety ``by design''.
This is both a timely as well as a necessary endeavour, because any
techniques that are to be used in practice will need to be
demonstrably safe, and, most importantly, will need to be accepted by
humans, i.e.\ they need to gain the trust of the humans who work with
them.


This paper is structured as follows.
Section~\ref{s:assurance} is focused on safety assurance; it presents
insights into existing and forthcoming standards covering the safety
of robotic co-workers in an industrial setting.
Section~\ref{s:lbd} investigates robotic learning from demonstration,
and we identify opportunities and challenges to integrate safety ``by
design''.
Section~\ref{s:hri} considers the psychological challenges associated
with human-robot co-worker teams arising in flexible manufacturing scenarios.
Finally, Section~\ref{s:conc} summarizes and concludes. 
%

\section{Assurance as a Foundation for Trust}\label{s:assurance}

\subsection{Existing and forthcoming standards for safety assurance}\label{s:existing}
The success of a human-robot co-worker team heavily relies on the
dependability of the robot.
Dependability has been defined as ``the ability to deliver a service
that can {\em justifiably} be trusted''~\cite{avizienis}.
Dependability is an over-arching concept that includes attributes such
as safety, availability, reliability, predictability, integrity and
maintainability.
Safety is a critical aspect of dependability, which traditionally is
assured prior to a system's deployment. Safety assurance assesses the
absence of harmful consequences of a system's actions (or in-action)
on users and the environment.

A key property of dependability assurance is that it is a subjective
condition of a system's users as well as an objective property of the
system itself. Even if a system design contains no flaws and its
operation never causes harm throughout its life, if its users cannot
be assured of this before they start to use the system, then they may
not trust the system and thus may never use it.
Therefore, the art of designing dependable systems is not only to
create a flawless design, but to do so in a manner that permits such
flawlessness to be demonstrated. This requires careful choices of a
system's architecture and mechanisms, because only those technologies
whose correct operation can be easily verified and validated are
suitable for such applications.

Until recently, the practical deployment of robotic assistants has
been held back by the lack of credible standards and techniques for
safety assurance. In traditional robotic applications, safety has been
achieved by confining robots to closed workplaces from which humans
are isolated.
Safety is assured by demonstrating that the isolation is effective,
rather than by demonstrating the safety of the technology of the
robot; the former is usually a much easier problem to solve.
Consequently, safety is not an essential aspect of such robots'
operations, and performance is the key objective.

Robotic co-workers, however, are designed to assist humans in their
work within the same workplace. It is inherent to their purpose that
they cannot be isolated from the people they are intended to assist.
Collaborative operation, however, in particular in a shared space, is a
much more complex problem to assure and requires more extensive
assessment of the technology of a robot and  its surroundings than
was necessary in more traditional applications.

Robots are expected to be safety-certified when entering
the market, which provides assurance that deploying a system does not
pose an unacceptable risk of adverse consequences.
%
%
Industry standards for collaborative operation in robots are currently
the focus of extensive study in the major international standards
agencies. The Technical Committee (TC)~184, Sub-Committee (SC)~2 of
the International Standards Organization (ISO) develops standards for
robots and robotic devices, and currently has working groups covering
industrial, medical, and service robots. Working Group (WG)~7 has
developed the ISO~13482 safety requirements standard for service
robots, including physical assistance and mobile servant applications,
and WG3 of the same committee is developing a Technical Specification
(TS)~15066 focused on collaborative robots.  TS~15066 provides guidance
on collaborative operation for industrial applications, including the
specification of several collaborative modes of operation and their
associated safety requirements.

Central to the guidelines in TS~15066 are hazard identification and
risk assessment. Both are performed by experts and are specific to the
collaborative task shared between a robot and its human operator.
Thus, task identification is key to the correct determination of any
foreseeable hazards. Risk assessment is then performed on these
hazards, including the identification of values such as the maximum
allowable speed of movement for the robot and the minimum separation
distance between robot and human, either as static values or as
dynamic ranges. Based on the risk assessment, risk reduction
strategies can be implemented for hazards where the risk of harm is
seen to be unacceptably high.

Because the traditional approach of risk reduction by separation of
the human from the robot cannot be used for collaborative settings,
the focus of the guidelines in TS~15066 is on influencing the design of
the robot, the joint workplace and the collaborative task itself, to
include protective measures that ensure the safety of human operators
at all times.
This may include re-design of tools and work pieces, e.g.\ to achieve
smooth, but not sharp, surfaces and low weights, both of which
influence the impact force in hazardous situations caused by direct
contact with humans.
%
%
To continuously track the position of humans within the collaborative
workplace, a speed and separation monitoring system is essential.
Where hazards arise out of direct contact with human operators,
whether intended or unintended, a fast contact detection system must
be in place to feed into a safety-related control system. This system
must be capable of processing context-related information in real time
and to activate protective measures when this becomes necessary. These
are examples of runtime monitors and an important part of safety
assurance at runtime, as we will see in the next section.

The impact force of dynamic contact and its duration, as well as the
location (body region) of such contact---all key parameters for safety
assessment---can vary greatly between collaborative tasks and, in
particular, from human to human, even when restricting human-robot
collaboration to shared workplaces within a flexible manufacturing
environment. This severely restricts the generality of calculations,
as, in principle, human-specific information is required.  It is still
to be determined how this problem of person-specific characterization
can be addressed. Potentially, a ``calibration'' phase may be required
before operation starts, so that the robot co-worker can be customized
to fit its human operator.


The ISO TS~15066 is expected to be publicly released later this
year.
It promotes a task-specific approach to safety assurance that requires
the hardware and software, the work environment and task specification
to be available for hazard analysis and risk assessment in their fully
finished forms prior to the deployment of the collaborative robotic
system. Furthermore, it is based on the assumption that it is possible
to predict as well as counter all hazardous operation conditions prior
to system deployment.
This appears to be inherently at odds with the concept of teaching
robots new tasks ``on the fly'', which leaves the task identification,
definition and training to the human operator, whose safety is of
paramount importance.  Hence, each newly learnt task, and also the
learning process itself, must be safety assured.
Thus, for any learning from demonstration technique to become viable
in practice, safety must be an integral part of the learning process,
directly embedded into the learning algorithms.  Task-specific safety
assurance, consequently, may need to be shifted to runtime, at least
in part.

\subsection{Assurance at runtime}

A case for ``Just-in-Time Certification'' of adaptive systems has been
made in~\cite{rushby:2007}.  Traditional assurance methods are based
on the assumption that, prior to deployment of a system, the system is
available in its final form for safety assessment, and
that all operating conditions that the system will face while
interacting with its environment can be predicted and analyzed
upfront. 
Adaptive systems, however, are designed to modify their behaviour at
runtime in response to changes in their environment or in the system
itself. 
Such systems simply do not meet the assumptions on which traditional
certification methods are based, because the time at which the
behaviour of the system is finalized is shifted from system design to 
system deployment. This impacts on the time at which certification can
be performed; it leaves at least part of the certification process to
be completed at runtime.
Adaptive systems, therefore, call for the development of novel
approaches to certification, not to replace traditional approaches,
but to complement these with techniques that can be used at runtime.

Robots that learn from demonstration are adaptive systems. They learn
new tasks at runtime. The fact that a human is involved in the
training as well as in the collaborative execution of the task is
associated with benefits and challenges with respect to safety
aspects.  On one side, the human operator can influence learning so
that the learning result complies with safety requirements. On the
other side, however, the human operator is exposed to potential
hazards during the learning process; this creates the obligation to
protect her/him from these. Can ``just-in-time'' techniques support
safety assurance in our context?

The provocatively named ``just-in-time'' certification approach
proposed in~\cite{rushby:2007} is firmly based on the use of formal
methods, at design time and at runtime.
It takes advantage of the observation that, if the behaviour of a
system was fixed at design time, then safety assessment would focus on
the pre-defined behaviour and determine whether its characteristics
meet a set of pre-defined safety requirements.
If the checking step could be automated, then its execution could
reasonably be shifted to runtime.
This, of course, necessitates encoding the safety requirements in a
suitable form for runtime monitoring, e.g.\ as a model or a protocol
to be adhered to.
A  monitor can then be formally derived from the model or
protocol.
At runtime, this monitor continuously checks compliance with the safety
requirements, and prevents any behaviour that causes violations.
 An advantage of performing these checks at runtime is that out of the
 huge number of possible system behaviours, only the one that is
 currently being adopted needs to be checked.
While, traditionally, such monitors are generated at design time and
applied at runtime, the use of Runtime Verification
techniques\footnote{\url{http://runtime-verification.org/}} enables
the generation of such monitors at runtime, based on explicit models
that capture the generic behaviour of system components and the safety
requirements that must be satisfied. 

In a similar way, ``learning from demonstration'' approaches could be
constrained at runtime, based on suitable models or protocols, to
deliver only learning results that are considered safe. Alternatively,
or in addition, task-specific safety monitors could be generated.
%
These prevent interactions which violate safety requirements at
runtime. It is worth noting that safety-related control systems are
already mentioned in the forthcoming standards for runtime monitoring
and potential intervention as briefly indicated in
Section~\ref{s:existing}.
In the next section we review ``learning from demonstration'' techniques
and investigate opportunities and challenges to integrate safety
assurance into the learning algorithms by design, as well as options
to shift part of safety assurance to runtime.


\section{Opportunities for Integrating Safety into the Learning from
  Demonstration Process}\label{s:lbd}

\subsection{Learning from Demonstration (LfD)}

To perform a task, a sequence of actions is applied to a given state
of the world. Each individual action transforms the world state; the
final action should result in a state that reflects the execution of
the task.
Finding an effective sequence of actions, i.e.\ a policy, to achieve a
target state is a key challenge in robotics and automation.
In ``Learning from Demonstration'' (LfD) robots learn a new task by
watching the task being performed by a human or a robot teacher.
%
%
The observations gained from watching the task guide a supervised
learning process towards the development of a policy that the robot
learner can use to perform the task.
LfD offers an intuitive way for humans to communicate with robots. It
enables non-experts to teach robots new skills by simply demonstrating
these to the robot.
In flexible manufacturing, robotic assistants are used to support
humans during the execution of a large variety of different tasks,
each of which would normally require some level of re-programming or
re-setting the robot.
With LfD techniques this is not necessary, as new tasks are acquired
by learning from example demonstrations. As such, LfD techniques
facilitate human-robot interaction and support more flexible
human-robot collaboration.

A variety of different machine learning techniques have been used for
LfD; a recent survey is contained in~\cite{argall}. In general, three
core approaches for policy derivation can be distinguished according
to~\cite{argall}:
\begin{enumerate}
\item[a)] those that learn directly how to map the robot's state
  observations to actions,
\item[b)] those that derive a policy based on learning a world model
  and a reward function, and
\item[c)] those that learn a policy by planning based on sequences of
  actions and their pre- and post-conditions.
\end{enumerate}
Irrespective of the learning approach, the challenge is to ensure that
safety is maintained during the learning process, and that the learning
result satisfies safety requirements ``by design''. This necessitates
embedding safety considerations directly into the learning algorithms.
How this can be achieved is a research question that needs to be
addressed before LfD can safely be deployed in practice.

The collaboration with robotic assistants is likely to change the
nature of the work, the tasks involved, and the work
environment~\cite{bradshaw:HART}.
For LfD it has been found that simple imitation or mimicking of the
demonstration is often not sufficient for the robot to perform the
task~\cite{atkeson}. Instead, the objective of the task that is being
demonstrated must be captured, so that a policy can be learnt that
enables the robot to achieve this objective.
The actual behaviour of the robot to realize its goal may differ from
that demonstrated, especially when the trainer is a human rather than
another robot. In~\cite{atkeson} it was found that the robot's hand
motion in a simple pendulum swing up task was quite different from the
motion recorded for the human demonstrator due to differences in
gripping technique and hand structure, which result in different task
dynamics.
In general, the bigger the physical differences between trainers and
learners, the more likely it is that the learnt behaviour differs,
although it achieves the same objective. 
This is an important finding and likely to be problematic in the
context of safety assurance, which is task specific, as described in
Section~\ref{s:existing}.
Even if the actions performed in a task demonstration satisfy safety
requirements, the safety assessment may need to be repeated for the
learning result, because the dynamics of the learnt task may differ
enough from that of the demonstration for the original safety
assessment to no longer hold.
%

The LfD approach 
in~\cite{atkeson} provides a good example to illustrate how safety may
be integrated into a model-based planning algorithm in the form of
constraints that capture pre-determined safety requirements.
%
The existing planner aims to find a policy that the robot can use to
accomplish the target task. Amongst all the possible policies, the
ones that  satisfy the relevant safety constraints are desirable.
While including safety constraints into planning clearly increases the
complexity of the learning task, 
this is necessary to ensure that  the learning outcome complies
with safety requirements.
In~\cite{atkeson} it has been shown that learning performance can be
significantly increased when background knowledge is provided in the
form of models that capture the physics of the task to be learnt. If
the physical characteristics of a task are known upfront, could safety
requirements be attached to these models to guide the learning towards
safe policies? Could these enriched models then serve as the basis for
the generation of runtime safety monitors?

\subsection{LfD with Reinforcement Learning}

Reinforcement Learning~\cite{sutton} is a widely used LfD
technique~\cite{argall} and a good example to illustrate the potential
problems arising when robots learn new tasks ``on the fly''.
The interesting feature of Reinforcement Learning is that policy
development is guided by feedback given to the robot during
explorative learning. Feedback is provided in the form of a function
that rewards desirable and penalizes undesirable actions. Learners aim
to maximise their cumulative reward. Based on the feedback, the robot
learns which actions or sequences of actions are preferable to achieve
a target goal.
Thus, in Reinforcement Learning, policy development necessitates
performing desirable as well as undesirable actions.
This can result in robots violating safety requirements while learning
new tasks. In an environment that requires close collaboration with
humans also during the learning stage, this cannot be tolerated.
Clearly, human operators need to be protected not only during the
joint task execution in collaboration with a robot, but also during
task transfer, i.e.\ during the LfD process.

An adaptation of Reinforcement Learning that includes future directed
rewards in the form of interactive guidance given by humans during the
learning process has been presented in~\cite{thomaz}.  The resulting
learning system can dynamically switch between explorative and
guidance-based learning. This has been achieved by modifying the
learning algorithm so that it accepts guidance when it is available;
if not, then the algorithm works by randomly selecting actions and
observing the associated reward.
Guidance is given by a human teacher who constrains the robot's action
selection, e.g.\ by focusing on a particular object, the explorative
learning is being restricted to the smaller and more relevant set of
actions related to that object.
The result is a more rational choice of actions compared to the random
action selection that would be made by algorithms without guidance. 
The benefits include, amongst others, improved learning performance
and a decreased number of failed trails.
This leads to more understandable behaviour of the learner and a more
fulfilling experience for the human teacher.
It seems plausible that a similar approach could be used to achieve
both, safety of the learning process and of the learning result, by
biasing action selection to those actions that satisfy safety
requirements. To achieve this, a method needs to be found to formalize
safety requirements as high-level policies that can guide learning.

\subsection{Virtual and mixed reality environments for LfD}

It has been proposed to transfer LfD into virtual environments to
reduce the time, labour and cost of real-world development
methods~\cite{sung}.  Virtual environments would also make the LfD
considerably safer for human operators.
In~\cite{sung}  modelling first creates a virtual agent, termed
the ``virtual human'', who takes the role of the human in LfD within a
virtual environment.  The virtual human is controlled by a human
operator to learn a human behavioural model which serves as a basis for
executing actions during the demonstration part of the learning.
Another virtual agent, termed the ``virtual learner'', observes the
virtual human's actions in order to learn behaviours that enable it to
collaborate with the human. This phase of the learning process is
called ``behavioural learning''.
In the second learning phase, the ``collaborative learning'', the
virtual learner then learns how to collaborate with the virtual human.
Finally, once learning has been accomplished in the virtual
environment, the results can be transferred to a real robot or a
software agent.

Safety compliance could reasonably be evaluated, at least in a first
instance, in such a virtual setting. Only those learning results that
have passed safety assessment in the virtual environment would then be
applied to the real robot. Of course, this necessitates a sufficiently
accurate model not only of the human and the robot learner, but also
of the environment in which the interaction will take place, i.e.\ the
collaborative workplace.
%
%
The major shortfall of the approach described in~\cite{sung} is that
it does not generalize to complex real world scenarios, where
interaction is required with different humans. As it stands, multiple
humans would need to be modelled and behaviour as well as
collaborative learning would need to be repeated for each human model.
The challenges of accommodating variability, both between
multiple demonstrations of a task by the same operator, and between
demonstrations of the same task by different operators, need to be
addressed to develop a more generic approach to LfD, not only in
virtual environments.

In~\cite{jakob} mixed-reality testbeds are used to support incremental
development of systems that involve humans, robots and software agents
with the goal to reduce costs and risks compared to testing in a
real-world environment.
The operation of these systems depends on a variety of factors,
including, but not limited to, those of the robot hardware, e.g.\ the
characteristics of the sensors and actuators, the features of the
environment, e.g.\ the collaborative workplace including tools and
materials, and, most importantly, the behaviour of the human
operators, resulting in a complex test environment overall.
To facilitate early and fast validation, the components in the
physical test environment can be replaced by virtual models in
incremental steps, either fully or partially.  The resulting
mixed-reality, multi-level testbed integrates models of different
fidelity and size, so that validation can be performed at the level of
abstraction that delivers the accuracy appropriate for the respective
test objective.

The degree of virtualization ranges from full virtualization of all
system components, as described in~\cite{sung}, to mixed-reality
settings where different system components are virtualized to
different degrees. The latter include hardware-in-the-loop simulation
environments to increase the fidelity of the hardware components in
the testbed, as well as human-in-the-loop virtual environments where
human operators work in a virtual reality setting so that the aspects
of human behaviour that are difficult (if not impossible) to capture
with virtual models become an integral part of the testbed.

In the context of human-robot collaborative manufacturing, the
fluctuations of human performance due to factors such as fatigue,
stress, lack of attention or concentration could thus be determined,
and their effects assessed, without exposing the human operator to the
risks inherent in testing these in the physical environment. Could
learning from demonstration be entirely shifted to virtual or mixed
reality environments to protect human operators during learning?

Many LfD techniques assume that for each action a well defined,
deterministic state transformation can be specified, and that all
system states are known and thus can be defined
upfront~\cite{argall}. This almost certainly does not hold for real
world settings, especially those involving humans. In the next section
we take a human-centric view on these aspects.

\section{Human Factors}\label{s:hri}

Humans have evolved as social animals over thousands of years. They
are highly sensitive to signals sent by other humans such as in
collaborative tasks~\cite{vesper:2010}. Any movement observed from
another agent (evolutionarily another human or an animal) will not
only be interpreted to have a purpose, but to be context-dependent and
meaningful~\cite{castelfranchi}.
For example, the speed with which a person hands over an object to
another person will vary depending on the physical abilities of the
two people involved in the interaction (e.g. an adult handing over a
cup to a small child or to another adult), the type of object (e.g. a
glass of water or a hot bowl of soup), and their intent (e.g. slow
movements might be a warning to the other person to be careful when
taking the object, but could also indicate reluctance to let go of
it). Therefore, humans will usually rely on the context they are in
and accompany their interactions with a range of (mostly nonverbal)
signals (such as eye gaze or facial expressions, body posture) with
the consequence of disambiguating their intentions.
Humans tend to anthropomorphise moving agents that are (seemingly)
able to adapt their behaviour in a dynamic way
and attribute to them social cognitive abilities of their
own~\cite{frith:2012}. This creates an intuitive expectation in the
human about the behaviour of such agents, which can be modulated only
by experience.

Imagine a very simple scenario in which a robot repeatedly performs a
single task which is highly predictable in its order of events, but
requires the robot to move within an arm's reach of the human (e.g.\
the robot picks up an object from a conveyor belt and holds it for the
human to work on). To make both the robot's and the human's tasks
efficient and to avoid collision between the human's arm or hand and
the robot itself, both human and robotic movements need to be exactly
orchestrated~\cite{valdesolo:2010}. This entails the implementation of
a cognitive model of human action in the robot that monitors and
predicts human action at a very short time scale, including human
inter- and intra-individual movement variability as well as cognitive
performance inconsistencies that are dependent on individual factors
such as fluid intelligence~\cite{ram:2009,ram:2005}. Measures of
intra-individual movement variability on a longer time scale such as
variability induced by fatigue are already widely used in Human
Factors (e.g.\ heart rate~\cite{rowe:1998,acharya:2006}, alpha
frequencies in electroencephalography to detect drowsiness~\cite{oken:2006,borghini:2012}.

Less common psychophysiological methods would be required to
provide the robot with dynamic information on a far shorter time
scale, such as tracking eye gaze to continuously monitor the human's
focus of attention~\cite{dehais:2012} and predict mind
wandering~\cite{forster}, or 3D motion capture information to track
the human's arm movements~\cite{isableu:2013}.
Models on Joint Action Understanding in humans~\cite{vesper:2010}
might give a first idea about the cognitive architecture necessary
within the robot.

Safety solutions to certification of specific robotic tasks for HRI as
given in the above scenario might seem complex and require a range of
control mechanisms, all regulating and controlling the human's
behaviour. However, they can, to large extents, rely on the inclusion
of already existing technology and methodology in Human Factors.

Imagine next a situation in which the robot selects between different
tasks depending on the context and is therefore not 100\% predictable
at all times.  This forces the human to deal with uncertainty about
the robot's next actions.
%
In such a scenario, it becomes even more important for the robot to
send effortlessly understandable signals that disambiguate its actions
for the human. This enables the human to maintain a sufficient level
of awareness of the state and actions of the robot. 
In fact, transparency together with control have been found to be more
important to human operators than increased
autonomy~\cite{johnson}. This is because, as the complexity of the
jointly performed task grows, so does the importance of addressing
mutual dependence to sustain team performance and the necessary level
of safety. Increased autonomy without addressing interdependence has
been shown to result in sub-optimal performance~\cite{johnson}.



\section{Summary and Conclusion}\label{s:conc}

Assurance methods have evolved over time, from isolation of humans
from robots, to a task-based approach in the forthcoming technical
specification, TS~15066.  There is, however, still a long way to go to
establish assurance methods for systems that acquire new behaviour at
runtime, such as robotic co-workers that learn from demonstration.

We propose that the question of safety needs to be addressed as an
integral part of the very process of teaching the robotic co-worker,
and have highlighted opportunities to extend existing LfD techniques
accordingly.
Furthermore, to achieve high levels of safety and ultimately trust,
the robotic co-worker must meet the innate expectations of the humans
it works with.
This necessitates equipping robotic co-workers with a cognitive model
of human action and the ability to clearly communicate their
intentions to human operators in a timely manner. 



\addtolength{\textheight}{-12cm}  

\section*{ACKNOWLEDGMENT}
The authors would like to thank Chris Melhuish, Anthony Pipe and
Dejanira Araiza Illan for fruitful discussions and feedback.
K Eder was partially supported by the UK EPSRC grant EP/K006320/1 ``Trustworthy Robotic
Assistants''.

\bibliographystyle{IEEEtran}
\bibliography{roman}

\end{document}